\documentclass[conference]{IEEEtran}
\IEEEoverridecommandlockouts
\usepackage{cite}
\usepackage{amsmath,amssymb,amsfonts}
\usepackage{algorithmic}
\usepackage{graphicx}
\usepackage{textcomp}
\usepackage{multicol} 
\usepackage{multirow}
\usepackage{booktabs}
\usepackage{hyperref}
\usepackage[table]{xcolor}
\usepackage{amssymb}

\definecolor{other flat1}{RGB}{175,0,75}
\newcommand{\hi}[1]{\textbf{\textcolor{other flat1}{#1}}}
\def\BibTeX{{\rm B\kern-.05em{\sc i\kern-.025em b}\kern-.08em
    T\kern-.1667em\lower.7ex\hbox{E}\kern-.125emX}}
\begin{document}

\title{DAGait: Generalized Skeleton-Guided Data Alignment for Gait Recognition
\thanks{\textsuperscript{*}Co-authors. \textsuperscript{\dag}Corresponding author.}}



\author{\IEEEauthorblockN{Zhengxian Wu\textsuperscript{*}}
\IEEEauthorblockA{\textit{ The Shenzhen International Graduate} \\
\textit{School, Tsinghua University}\\
Shenzhen, China \\
zx-wu24@mails.tsinghua.edu.cn}
\and
\IEEEauthorblockN{Chuanrui Zhang\textsuperscript{*}}
\IEEEauthorblockA{\textit{ The Shenzhen International Graduate} \\
\textit{School, Tsinghua University}\\
Shenzhen, China \\
zhang-cr22@mails.tsinghua.edu.cn}
\and
\IEEEauthorblockN{Hangrui Xu}
\IEEEauthorblockA{\textit{ School of Computer Science and } \\
\textit{Information Engineering,}\\
\textit{Hefei University of Technology}\\
Hefei, China \\
2022217415@mails.hfut.edu.cn}
\and
\hspace*{3cm}
\IEEEauthorblockN{Peng Jiao}
\IEEEauthorblockA{\hspace*{3cm}\textit{ The Shenzhen International Graduate} \\
\hspace*{3cm}\textit{School, Tsinghua University}\\
\hspace*{3cm}Shenzhen, China \\
\hspace*{3cm}jiaop21@mails.tsinghua.edu.cn}
\and
\IEEEauthorblockN{Haoqian Wang\textsuperscript{\dag}}
\IEEEauthorblockA{\textit{ The Shenzhen International Graduate} \\
\textit{School, Tsinghua University}\\
Shenzhen, China \\
wanghaoqian@tsinghua.edu.cn}
}

\maketitle
\begin{abstract}
Gait recognition is emerging as a promising and innovative area within the field of computer vision, widely applied to remote person identification. 
Although existing gait recognition methods have achieved substantial success in controlled laboratory datasets, their performance often declines significantly when transitioning to wild datasets.
We argue that the performance gap can be primarily attributed to the spatio-temporal distribution inconsistencies present in wild datasets, where subjects appear at varying angles, positions, and distances across the frames. 
To achieve accurate gait recognition in the wild, we propose a skeleton-guided silhouette alignment strategy, which uses prior knowledge of the skeletons to perform affine transformations on the corresponding silhouettes.
To the best of our knowledge, this is the first study to explore the impact of data alignment on gait recognition. 
We conducted extensive experiments across multiple datasets and network architectures, and the results demonstrate the significant advantages of our proposed alignment strategy.
Specifically, on the challenging Gait3D dataset, our method achieved an average performance improvement of 7.9\% across all evaluated networks. 
Furthermore, our method achieves substantial improvements on cross-domain datasets, with accuracy improvements of up to 24.0\%.Code is available at: \url{https://github.com/DingWu1021/DAGait}
\end{abstract}


\section{Introduction}
\label{sec:intro}
Gait recognition identifies individuals by analyzing the unique movement patterns associated with human walking. 
Due to its non-intrusive and distinctive characteristics, it has emerged as a promising biometric modality. 
In comparison to other biometric methods, such as facial recognition \cite{face-recognition} and fingerprint identification \cite{fingerprint-recognition}, gait recognition offers the advantages of non-contact operation and is hard to disguise.
These features make gait recognition particularly advantageous in applications such as identity authentication , security surveillance , and public safety \cite{ID-recognition,muaaz2017smartphone,secure}.

\begin{figure}[t]
    \begin{center}
    \includegraphics[width=1.0\linewidth]{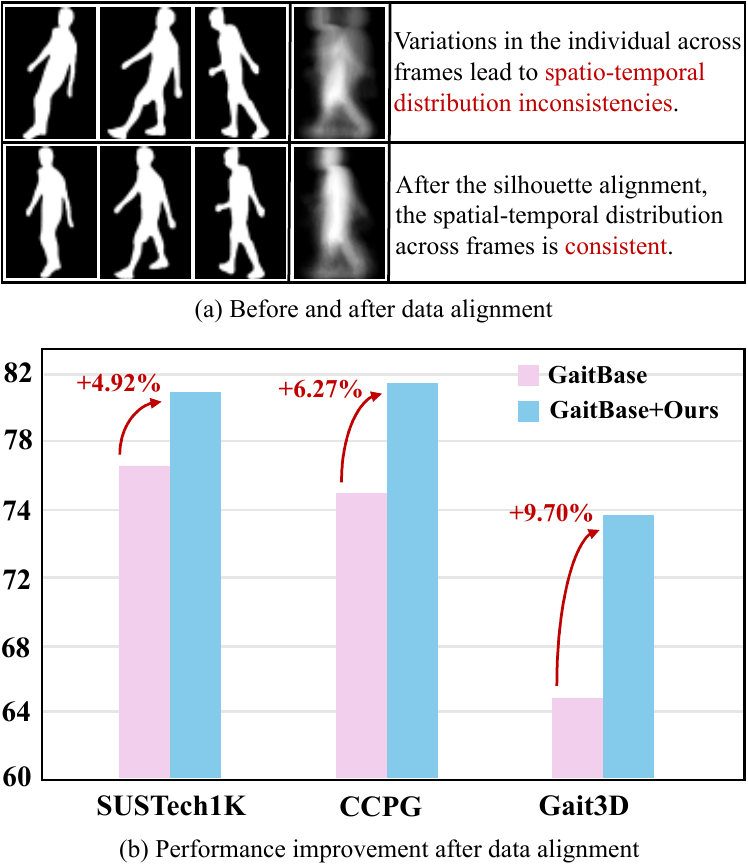}
    \end{center}
    \vspace{-4.0 mm}
    \caption{(a) The figure demonstrates the silhouette and gait energy image (GEI) within a sequence before and after data alignment, highlighting the alignment's effectiveness in mitigating spatio-temporal distribution inconsistencies. (b) The performance comparison of GaitBase\cite{OpenGait} without and with data alignment across various datasets, demonstrating significant accuracy improvements, particularly on the Gait3D\cite{Gait3D} wild dataset.}
    \label{fig1}
\end{figure}

Binary silhouettes are widely used in gait recognition due to their effectiveness in mitigating background noise. 
Early deep learning approaches that use silhouette sequences as input \cite{xulie1, xulie2, GaitPart} have shown promising results in controlled laboratory datasets, such as CASIA-B \cite{CASIA-B} and OU-MVLP \cite{OU-MVLP}. 
These methods typically rely on local convolutional operations and pooling layers to extract spatial and temporal features.
However, when the scene is changed to wild datasets, such as GREW\cite{Grew} and Gait3D\cite{Gait3D}, the performance of these methods often decline significantly.
We argued that this degradation is primarily caused by spatio-temporal distribution inconsistencies in wild datasets.
Specifically,variations in factors such as camera angles, human poses, and occlusions lead to significant differences in subjects within a single sequence.
Due to these inconsistencies, the same areas across frames may represent different features, leading to confusion in the local feature extraction and spatio-temporal pooling. 

Recent methods typically employ general data augmentation strategies and more complex network architectures to improve model recognition performance in real-world scenarios\cite{deepgait,visual_gait,ShinanZouMSAFF}. 
We observe that these methods primarily focus on capturing the differences in human posture distributions and the movement trends of key body parts through the design of complex network models. 
However, complex network architectures are prone to overfitting noise and local patterns. 
Moreover, these deep models often lack generalization, as they tend to learn specific details from the training data rather than learning robust features. 
In contrast, data alignment is a more effective strategy.
Accurate data alignment can mitigate interference from perspective shifts and posture variations, enabling the network to learn universal gait features.
Unfortunately, this aspect is often overlooked. Therefore, we argue that relying solely on general data augmentation strategies is insufficient to address the challenges posed by wild datasets.


Based on the above analyses, we propose a gait recognition framework, named \textbf{DAGait}, designed to address the distribution discrepancies commonly observed in wild datasets. 
Specifically, we propose a skeleton-guided silhouette alignment strategy that utilizes the spatial relationships between skeleton joints and corresponding silhouette regions to apply affine transformations for correction. 
This strategy ensures that the pose in each frame is aligned perpendicular to the ground and scaled uniformly. 
Our proposed strategy enables each individual to obtain gait features with consistent spatio-temporal distribution. 
Consequently, as shown in Fig.~\ref{fig1}, our approach significantly improves the accuracy of existing methods on the Gait3D \cite{Gait3D}, CCPG \cite{CCPG}, and SuTech-1K \cite{SUSTech1K} datasets. 
Furthermore, data augmentation presents a strong cross-dataset generalization ability.
In cross-dataset experiments, with CCPG and Gait3D serving as the training and testing sets, it achieved a 24\% accuracy improvement.

Our main contributions are summarized as follows:

\vspace{-0.5mm}
\begin{itemize}
\item We discuss the importance of data alignment for gait recognition task in real-world scenarios and propose a skeleton-based data alignment strategy to address inconsistencies in spatio-temporal distribution.
\item We introduce a novel gait recognition framework, \textbf{DAGait}, which utilizes spatio-temporally consistent gait data and achieves notable performance improvements across various gait recognition models.
\item Our approach significantly improves the accuracy of existing methods across various datasets, with the highest improvement of up to 15.7\% on the wild dataset Gait3D and 24.0\% on cross-dataset evaluations.

\end{itemize}

\section{RELATED WORK}

\textbf{Gait Recognition.} Gait recognition methods can be classified into two main categories: model-based and appearance-based methods.
Model-based methods consider the basic physical structure of the human body \cite{Human-struction} and use interpretable models to represent gait characteristics. 
These methods recognize individuals by extracting human posture information \cite{Human-pose, Human-pose2} to model body structure and walking patterns. 
However, model-based approaches are less effective than model-free methods because they lack valuable gait information such as shape and appearance.
On the other hand, appearance-based methods \cite{appearance1} utilize binarized human silhouette images\cite{sil} to capture inherent spatial and temporal variations in body shape, clothing, and movement dynamics.
Specifically, GaitSet\cite{gaitset}innovatively treats the gait sequence as a set and employs a maximum function to compress the sequence of frame-level spatial features.
GaitGL\cite{gaitgl} argues that spatially global gait representations often overlook critical details, while local region-based descriptors fail to capture relationships between neighboring parts.

\textbf{Data augmentation strategy.} Existing methods typically adopt a unified data augmentation strategy\cite{OpenGait} for network input, including horizontal flipping, rotation, and random perspective transformation. 
Fu et al.\cite{GPgait} proposed a novel skeleton enhancement strategy to address the generalization challenges in skeleton-based recognition. 
Wang et al.\cite{QAGait} introduced a quality evaluation strategy and perceptual feature learning method, focusing on dataset quality. In contrast, our work aims to investigate the significance of data alignment in gait recognition.

\begin{figure*}[ht]
    \begin{center}        \includegraphics[width=1.0\linewidth]{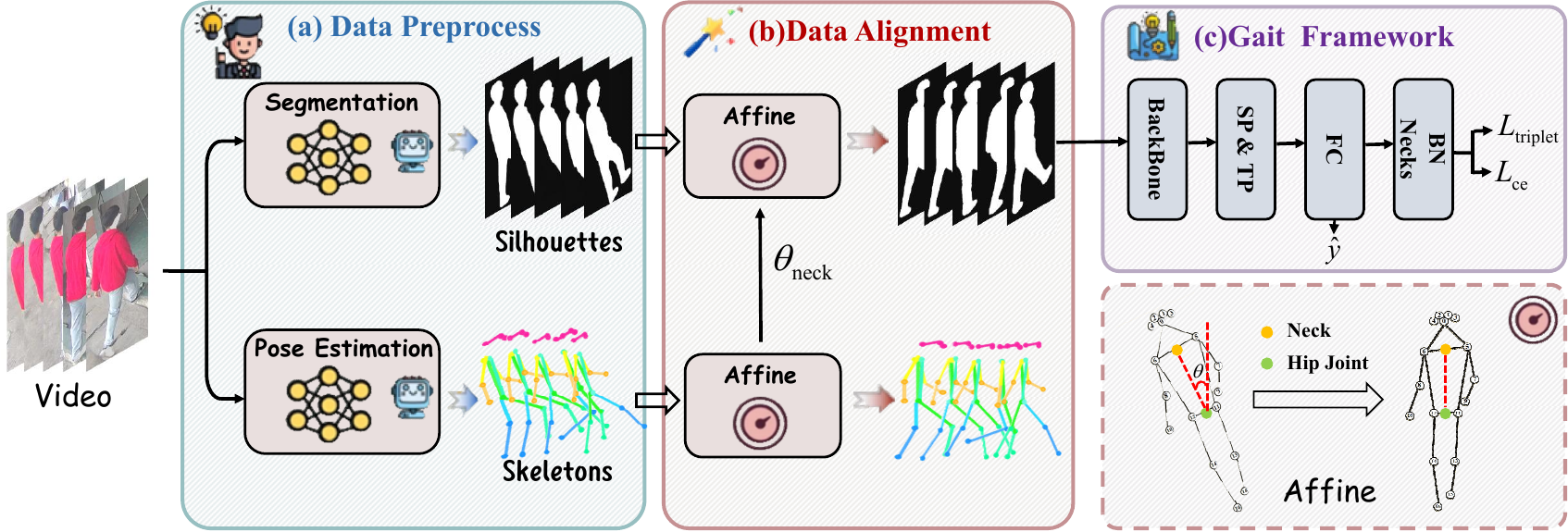}
    \end{center}
    \caption{Overview of our proposed data-alignment-based gait recognition
    framework \textbf{DAGait}. In the data preprocessing stage, silhouettes and skeletons are extracted from the RGB image sequences. During the data alignment phase, prior spatial relationships between skeleton joints and silhouette regions are used to apply an affine transformation, correcting the silhouette. Finally, the aligned silhouette images are input into the backbone network for recognition.}
        \label{fig2}
\end{figure*} 

\section{METHOD}
This section provides a comprehensive explanation of the data alignment process. 
In Sec.~\ref{A}, we present the overall architecture and data processing pipeline for gait recognition. 
Sec.~\ref{B} focuses on the affine transformation of skeleton maps. In Sec.~\ref{C}, we propose a silhouette alignment strategy guided by the skeleton. 
Finally, Sec.~\ref{D} discusses the application of data augmentation to mitigate overfitting and improve the model's generalization ability.

\subsection{Gait Recognition Pipeline}\label{A}

Fig.~\ref{fig2} illustrates the overall architecture of the gait recognition framework \textbf{DAGait}. 
Silhouette-based gait recognition methods typically involve two critical stages: data preprocessing and data augmentation, performed before the data is fed into the network.  
In the data preprocessing stage, the silhouette is first cropped to exclude regions above the head and below the feet.
The image is then resized to a uniform height and horizontally shifted to the center. 
Finally, the redundant side areas are cropped to achieve the specified width. 
Subsequently, data augmentation is employed to increase the diversity of the training dataset by implementing operations such as horizontal flipping, random affine transformations and random erasure. These methods enhance the model’s generalization capabilities and reduce the risk of overfitting.

However, in wild datasets, significant variations in motion trajectories, viewpoints, and the relative positions of subjects across frames often render traditional preprocessing and alignment methods inadequate for addressing these inconsistencies.
Additionally, these discrepancies can be further intensified during the data augmentation process, negatively affecting the network's recognition accuracy.
Consequently, we argue that to achieve better performance in wild datasets, it is essential to adopt additional data alignment strategies to reduce the distributional discrepancies.

\subsection{Alignment of Skeleton Diagram}\label{B}

Previous studies \cite{GPgait} have observed that skeleton maps exhibit spatio-temporal inconsistencies and can be addressed through affine transformation. 
Building on this, we propose that affine transformations to skeletons offer valuable prior information for aligning silhouettes.

In our approach, the skeleton sequence of the identified individual is denoted as \( X_{ske} \in \mathbb{R}^{N \times 3 \times Z} \). 
The neck position, \( p_{neck} \) is defined as the midpoint between the left and right shoulders, while the hip position  \( p_{hip} \) is defined as the midpoint between the left and right hips.
The line connecting the neck and the hip is regarded as the spine. 
We hypothesize that during walking the spine should remain perpendicular to the ground. 
Therefore, we apply an affine transformation, using the neck as the center of rotation, to rotate the spine ensuring it remains consistently vertical to the ground.
The rotation angle \(\theta \) is computed using the following formula:
\begin{eqnarray}
\theta = \mathrm{\arctan}\left(\frac{{x}_{neck}-{x}_{hip}}{{y}_{neck}-{y}_{hip}+\epsilon}\right).\label{1}
\end{eqnarray}
Where \( (x_{neck}, y_{neck})\) and \( (x_{hip}, y_{hip})\) represent the coordinates of the neck and hip, and \(\epsilon\)  is the residual factor, introduced to prevent division by zero.
For each frame in the gait sequence, we apply the affine transformation to rotate the spine ensuring it is aligned vertically with the ground:
\begin{eqnarray}
\left[
\begin{array}{c}
    x^{\prime} \\
    y^{\prime}
\end{array}\right] = R \left[
\begin{array}{c}
    x \\
    y
\end{array}\right] + T, \label{2}
\end{eqnarray}
\begin{eqnarray}
R = \left[
\begin{array}{cc}
    \cos\theta & -\sin\theta \\
    \sin\theta & \cos\theta
\end{array}\right], \label{3}
\end{eqnarray}

\begin{eqnarray}
T = \left[
\begin{array}{c}
    (1 - \cos\theta) \cdot x_{neck} + \sin\theta \cdot y_{neck} \\
    (1 - \cos\theta) \cdot y_{neck} - \sin\theta \cdot x_{neck}
\end{array}\right]. \label{4}
\end{eqnarray}
The parameters $(x, y)$ and $(x', y')$ represent the coordinates of a point before and after the transformation, while $ \theta $ represents the rotation angle. 
In this module, we obtain the rotation angle $ \theta $ and the neck position coordinates, which are subsequently used for aligning the silhouette map.

\subsection{Skeleton-guided Silhouette Alignment}\label{C}
The primary goal of silhouette map alignment is to minimize spatio-temporal inconsistencies caused by various factors. 
In our approach, the silhouette sequence of the identified individual is denoted as  \( X_{sil} \in \mathbb{R}^{N \times H \times W} \) and it is aligned with the skeleton. 
For each frame, we utilize the neck position \( (x_{neck},y_{neck})\) and the rotation angle $ \theta $ from the skeleton as the reference origin and rotation angle for transforming the corresponding silhouette map. 
This allows us to generate an affine transformation matrix that precisely adjusts the pose of the silhouette map. 
The affine transformation matrix is expressed as:


\begin{equation}
M = \left[
\begin{array}{l|l}
    R & T
\end{array}\right].
\end{equation}
This matrix is then applied to each frame of the silhouette sequence to perform rotation and translation:
\begin{eqnarray}
\left[
\begin{array}{llllllllll}
	{x}^{^{\prime}} \\
	{y}^{^{\prime}}
\end{array}\right]=M\cdot\left[
\begin{array}{llllllllll}
	x \\
	y
\end{array}\right].\label{6}
\end{eqnarray}
The parameters $(x, y)$ and $(x', y')$ represent the coordinates of a point before and after the transformation. 
This transformation ensures the initial alignment of the subject’s pose; additionally, scaling and displacement adjustments are required to ensure spatial consistency throughout the sequence. 
Specifically, we scale the foreground region of the silhouette to a uniform size and align the frames using the neck position as the reference point. 
This additional alignment ensures that the silhouettes are consistently positioned across all frames, effectively obtaining gait features with spatio-temporal distribution consistency.

\begin{table*}[t]
    \footnotesize
    \centering
    \caption{Performance comparison of the original Gait3D, CCPG, and SUSTech1K datasets versus their data-aligned counterparts, evaluated across multiple gait recognition networks.}        
    \label{tab1}
    \begin{tabular}{cl|cccccccccc}
        \toprule
            \multicolumn{2}{c|}{\multirow{4}{*}{Method}}  & \multicolumn{10}{c}{Testing Datasets}                                                                                                                                                                     \\ \cline{3-12} 
            \multicolumn{2}{c|}{}                         & \multicolumn{3}{c|}{Gait3D}                                                                   & \multicolumn{5}{c|}{CCPG}                             & \multicolumn{2}{c}{SUSTech1K}                     \\ \cline{3-12} 
            \multicolumn{2}{c|}{}                         & \multirow{2}{*}{Rank-1} & \multirow{2}{*}{Rank-5} & \multicolumn{1}{c|}{\multirow{2}{*}{mAP}} & CL   & UP   & DN   & BG   & \multicolumn{1}{c|}{Mean} & \multirow{2}{*}{Rank-1} & \multirow{2}{*}{Rank-5} \\ \cline{6-10}
            \multicolumn{2}{c|}{}                         &                         &                         & \multicolumn{1}{c|}{}                     & \multicolumn{5}{c|}{Rank-1}                           &                         &                         \\ \hline
            \multicolumn{2}{c|}{GaitPart(CVPR2020)\cite{GaitPart}}       & 28.2                    & 47.6                    & \multicolumn{1}{c|}{21.6}                 & -    & -    & -    & -    & \multicolumn{1}{c|}{-}    & 59.2                    & 80.7                    \\
            \rowcolor{gray!10} \multicolumn{2}{c|}{GaitPart+Ours}                     & 43.9$\mathrm{\tiny{\hi{+15.7}}}$                    & 64.9$\mathrm{\tiny{\hi{+17.3}}}$                    & \multicolumn{1}{c|}{35.0$\mathrm{\tiny{\hi{+13.4}}}$}                 & -    & -    & -    & -    & \multicolumn{1}{c|}{-}    & 61.5$\mathrm{\tiny{\hi{+2.3}}}$                    & 81.4$\mathrm{\tiny{\hi{+0.7}}}$                    \\
            \multicolumn{2}{c|}{GaitSet(TPAMI2021)\cite{gaitset}}       & 36.7                    & 58.3                    & \multicolumn{1}{c|}{30.0}                 & -    & -    & -    & -    & \multicolumn{1}{c|}{-}    & 65.0                    & 84.7                    \\
            \rowcolor{gray!10} \multicolumn{2}{c|}{GaitSet+Ours}                     & 49.3$\mathrm{\tiny{\hi{+12.6}}}$                    & 67.8$\mathrm{\tiny{\hi{+9.5}}}$                    & \multicolumn{1}{c|}{38.0$\mathrm{\tiny{\hi{+8.0}}}$}                 & -    & -    & -    & -    & \multicolumn{1}{c|}{-}    & 66.2$\mathrm{\tiny{\hi{+1.2}}}$                    & 84.5$\mathrm{\tiny{\hi{-0.2}}}$                    \\
            \multicolumn{2}{c|}{GaitBase(CVPR2023)\cite{OpenGait}}       & 64.6                    & 80.0                    & \multicolumn{1}{c|}{54.5}                 & 71.9 & 75.1 & 76.7 & 78.5 & \multicolumn{1}{c|}{75.5} & 76.1                    & 89.3                    \\
            \rowcolor{gray!10} \multicolumn{2}{c|}{GaitBase+Ours}                     & 74.3$\mathrm{\tiny{\hi{+9.7}}}$                    & 87.9$\mathrm{\tiny{\hi{+7.9}}}$                    & \multicolumn{1}{c|}{66.2$\mathrm{\tiny{\hi{+11.7}}}$}                 & 77.0$\mathrm{\tiny{\hi{+5.1}}}$ & 81.1$\mathrm{\tiny{\hi{+6.0}}}$ & 82.5$\mathrm{\tiny{\hi{+5.8}}}$ & 86.5$\mathrm{\tiny{\hi{+8.0}}}$ & \multicolumn{1}{c|}{81.8$\mathrm{\tiny{\hi{+6.3}}}$} & 77.6$\mathrm{\tiny{\hi{+1.5}}}$                    & 90.5$\mathrm{\tiny{\hi{+1.2}}}$                    \\ \hline
            \multicolumn{2}{c|}{DGaitv2-2D-L(Arxiv2023)\cite{deepgait}}  & 67.8                    & 83.9                    & \multicolumn{1}{c|}{59.7}                 & 69.9 & 76.1 & 77.1 & 84.4 & \multicolumn{1}{c|}{76.9} & 74.8                    & 89.2                    \\
            \rowcolor{gray!10} \multicolumn{2}{c|}{DGaitv2-2D-L+Ours}                     & 75.2$\mathrm{\tiny{\hi{+7.4}}}$                    & 88.4$\mathrm{\tiny{\hi{+4.5}}}$                    & \multicolumn{1}{c|}{67.2$\mathrm{\tiny{\hi{+7.5}}}$}                 & 73.8$\mathrm{\tiny{\hi{+3.9}}}$ & 78.0$\mathrm{\tiny{\hi{+1.9}}}$ & 81.4$\mathrm{\tiny{\hi{+4.3}}}$ & 86.2$\mathrm{\tiny{\hi{+1.8}}}$ & \multicolumn{1}{c|}{80.0$\mathrm{\tiny{\hi{+3.1}}}$} & 77.3$\mathrm{\tiny{\hi{+2.5}}}$                    & 90.2$\mathrm{\tiny{\hi{+1.0}}}$                    \\
            \multicolumn{2}{c|}{DGaitv2-3D-L(Arxiv2023)\cite{deepgait}}  & 74.1                    & 87.0                    & \multicolumn{1}{c|}{66.5}                 & 77.9 & 83.2 & 79.9 & 88.6 & \multicolumn{1}{c|}{82.4} & 79.4                    & 91.2                    \\
            \rowcolor{gray!10} \multicolumn{2}{c|}{DGaitv2-3D-L+Ours}                     & 78.5$\mathrm{\tiny{\hi{+4.4}}}$                    & 89.7$\mathrm{\tiny{\hi{+2.7}}}$                    & \multicolumn{1}{c|}{71.0$\mathrm{\tiny{\hi{+4.5}}}$}                 & 79.6$\mathrm{\tiny{\hi{+1.7}}}$ & 85.0$\mathrm{\tiny{\hi{+1.8}}}$ & \textbf{84.1}$\mathrm{\tiny{\hi{+4.2}}}$ & 88.7$\mathrm{\tiny{\hi{+0.1}}}$ & \multicolumn{1}{c|}{84.4$\mathrm{\tiny{\hi{+2.0}}}$} & 81.5$\mathrm{\tiny{\hi{+2.1}}}$                    & 92.1$\mathrm{\tiny{\hi{+0.9}}}$                    \\
            \multicolumn{2}{c|}{DGaitv2-3D-H(Arxiv2023)\cite{deepgait}}  & 75.8                    & 87.3                       & \multicolumn{1}{c|}{67.3}                    & 75.9 & 83.2 & 79.1 & 88.7 & \multicolumn{1}{c|}{81.8} & 80.4                    & 91.6                    \\
            \rowcolor{gray!10} \multicolumn{2}{c|}{DGaitv2-3D-H+Ours}                     & 79.4$\mathrm{\tiny{\hi{+3.6}}}$                    & 89.8$\mathrm{\tiny{\hi{+2.5}}}$                    & \multicolumn{1}{c|}{72.7$\mathrm{\tiny{\hi{+5.4}}}$}                 & 78.0$\mathrm{\tiny{\hi{+2.1}}}$ & 83.8$\mathrm{\tiny{\hi{+0.6}}}$ & 83.2$\mathrm{\tiny{\hi{+4.1}}}$ & 89.0$\mathrm{\tiny{\hi{+0.3}}}$ & \multicolumn{1}{c|}{83.5$\mathrm{\tiny{\hi{+1.7}}}$} & \textbf{82.8}$\mathrm{\tiny{\hi{+2.4}}}$                    & 92.7$\mathrm{\tiny{\hi{+1.1}}}$                    \\
            \multicolumn{2}{c|}{DGaitv2-P3D-L(Arxiv2023)\cite{deepgait}} & 74.2                    & 86.9                    & \multicolumn{1}{c|}{67.1}                 & 76.8 & 82.5 & 79.8 & 88.2 & \multicolumn{1}{c|}{81.8} & 79.6                    & 91.4                    \\
            \rowcolor{gray!10} \multicolumn{2}{c|}{DGaitv2-P3D-L+Ours}                     & 78.9$\mathrm{\tiny{\hi{+4.7}}}$                    & 89.8$\mathrm{\tiny{\hi{+2.9}}}$                    & \multicolumn{1}{c|}{71.5$\mathrm{\tiny{\hi{+4.4}}}$}                 & 79.9$\mathrm{\tiny{\hi{+3.1}}}$ & 84.0$\mathrm{\tiny{\hi{+1.5}}}$ & 83.8$\mathrm{\tiny{\hi{+4.0}}}$ & 89.0$\mathrm{\tiny{\hi{+0.8}}}$ & \multicolumn{1}{c|}{84.3$\mathrm{\tiny{\hi{+2.5}}}$} & 81.6$\mathrm{\tiny{\hi{+2.0}}}$                    & 92.3$\mathrm{\tiny{\hi{+0.9}}}$                    \\
            \multicolumn{2}{c|}{DGaitv2-P3D-H(Arxiv2023)\cite{deepgait}} & 75.0                    & 87.3                       & \multicolumn{1}{c|}{66.9}                    & 76.2 & 83.5 & 80.0 & 89.4 & \multicolumn{1}{c|}{82.3} & 80.7                    & 91.7                    \\
            \rowcolor{gray!10} \multicolumn{2}{c|}{DGaitv2-P3D-H+Ours}                     & \textbf{80.4}$\mathrm{\tiny{\hi{+5.4}}}$                    & \textbf{90.4}$\mathrm{\tiny{\hi{+3.1}}}$                    & \multicolumn{1}{c|}{\textbf{72.9}$\mathrm{\tiny{\hi{+6.0}}}$}                 & \textbf{80.5}$\mathrm{\tiny{\hi{+4.3}}}$ & \textbf{86.0}$\mathrm{\tiny{\hi{+2.5}}}$ & 83.8$\mathrm{\tiny{\hi{+3.8}}}$ & \textbf{89.6}$\mathrm{\tiny{\hi{+0.2}}}$ & \multicolumn{1}{c|}{\textbf{85.0}$\mathrm{\tiny{\hi{+2.7}}}$} & 82.6$\mathrm{\tiny{\hi{+1.9}}}$                    & \textbf{92.9}$\mathrm{\tiny{\hi{+1.2}}}$                    \\         

        \bottomrule 
    \end{tabular}
\end{table*}

\begin{table}[t]
    \caption{Comparison of Gait3D performance before and after alignment on multimodal networks.}
    \label{tab2}
    \centering
    \begin{tabular}{c|cccc}
        \toprule
        \multirow{2}{*}{Method}  & \multicolumn{4}{c}{Gait3D}    \\ \cline{2-5} 
                                 & Rank-1 & Rank-5 & mAP  & mINP \\ \hline
        MSAFF(IJCB2023)\cite{ShinanZouMSAFF}          & 48.1   & 66.6   & 38.4 & 23.5 \\
        \rowcolor{gray!10}MSAFF+ours               & 52.6$\mathrm{\tiny{\hi{+4.5}}}$    & 73.7$\mathrm{\tiny{\hi{+7.1}}}$    & 42.3$\mathrm{\tiny{\hi{+3.9}}}$  & 28.8$\mathrm{\tiny{\hi{+5.3}}}$  \\
        SkeletonGait(AAAI2024)\cite{skeletongaitgaitrecognitionusing} & 77.6   & 89.4   & 70.3 & 42.6 \\
        \rowcolor{gray!10}SkeletonGait+Ours      & 79.8$\mathrm{\tiny{\hi{+2.2}}}$   & 90.5$\mathrm{\tiny{\hi{+1.1}}}$   & 72.6$\mathrm{\tiny{\hi{+2.3}}}$ & 50.7$\mathrm{\tiny{\hi{+8.1}}}$ \\
        \bottomrule 
    \end{tabular}
\end{table}

\subsection{Data Augmentation}\label{D}
To enhance the model's generalization ability and effectively mitigate the risk of overfitting, data augmentation is incorporated during the training process. 
Given that the frames in the sequence are already spatially aligned, we adopt a general image data augmentation strategy \cite{OpenGait}, which includes the following methods:
(1)\textbf{Horizontal Flipping.} By flipping the image horizontally, the model is exposed to more diverse walking directions, thereby enhancing its robustness to variations in walking orientation. 
This transformation does not affect the gait features, while simulating different walking directions in real-world scenarios.
(2)\textbf{Affine and Perspective Transformations.} This operation simulates variations in viewpoint by applying random affine and perspective transformations to the image. 
Although the key features are aligned in the spatial and temporal dimensions, small angular transformations can help improve the model’s robustness to subtle viewpoint changes, allowing it to better generalize across different environments.
(3)\textbf{Random Erasure.} The random Erasure method randomly occludes certain parts of the image to simulate occlusion or background noise commonly found in real-world scenarios. 
Random erasure encourages the model to focus on global gait features, reducing the model’s tendency to overfit local or irrelevant details.

These data augmentation strategies enable the model to learn richer training samples, enhancing its adaptability and robustness to varying environmental conditions.

\section{EXPERIMENTS}
\subsection{Datasets}
In this study, our method is evaluated on three challenging gait datasets, including the wild dataset \textbf{Gait3D} \cite{Gait3D}, and two cross-garment and multi-view datasets \textbf{CCPG} \cite{CCPG} and \textbf{SUSTech1K} \cite{SUSTech1K}. 
Among these, the \textbf{Gait3D} dataset serves as the primary dataset for evaluation due to its highly representative complex real-world scenario.

\textbf{Gait3D} was captured in a large supermarket environment, using 39 high-resolution cameras (1920 × 1080 resolution, 25 FPS) to record a total of 1,090 hours of video footage. 
After preprocessing, the dataset contains gait data from 4,000 subjects, including 25,309 gait sequences and more than 3,279,239 frames. 
The dataset presents a challenging setting with a highly complex environmental background, which includes variations in perspective, occlusions, and other environmental interferences during the gait capture process. 
These characteristics make Gait3D particularly suitable for evaluating the effectiveness and robustness of data alignment methods in real-world, variable environments.

\textbf{CCPG} is a gait recognition dataset specifically designed for addressing challenges caused by clothing variations.
It contains 200 subjects wearing many different clothes and over 16,000 sequences, and the RGB data is available. 
\textbf{SUSTech1K} is collected by a LiDAR sensor and an RGB camera. 
The dataset contains 25,239 sequences from 1,050 subjects and covers many variations, including visibility, views, occlusions, clothing, carrying, and scenes.
The dataset captures data streams from LiDAR and camera sensors, opening up opportunities for exploring sensor fusion approaches for robust gait recognition.
\subsection{Implementation Details}
The experimental implementation in this study followed the official protocols for each dataset, including the standard partitioning strategies for the training, gallery, and probe sets. 
we measure its distance between all sequences in the gallery set. Then a ranking list of the gallery set is returned by the ascending order of the distance. We adopt the average Rank-1, Rank-5, and mean Average Precision (mAP) as the evaluation metrics.
During testing, a comprehensive gait evaluation protocol for multi-view scenarios was applied across all datasets, with Rank-1 accuracy serving as the primary evaluation metric. 
Regarding dataset resolution, the image resolution for Gait3D is set to (64, 44), while both CCPG and SUSTech1K datasets uniformly utilize a resolution of (64, 64).

To ensure fairness across all experiments, a consistent data augmentation strategy was applied. The data augmentation techniques include random horizontal flipping, random rotations, and random occlusions, each with a probability of 20\%. Additionally, we standardized the training parameters, including the number of training epochs, learning rate, weight decay, and optimizer settings, following the recommendations provided by the OpenGait\cite{OpenGait} framework.

\subsection{Comparison with State-of-the-Art (SOTA)}
As shown in Table~\ref{tab1}, the proposed data alignment strategy consistently improves the performance of existing methods across all three datasets, with particularly significant effects observed on the more challenging wild dataset, Gait3D.

\noindent\textbf{Evaluation of the Gait3D Dataset.} The proposed data alignment strategy effectively reduces inconsistencies within the Gait3D dataset.
Specifically, after alignment, the Rank-1, Rank-5, and mAP of each network showed average increases of 7.9\%, 5.8\%, and 7.3\%, respectively. 
Furthermore, on GaitPart \cite{GaitPart}, the Rank-1 accuracy improved by 15.7\%, corresponding to a 55.6\%  relative increase from the original result (43.9\% vs. 28.2\%). 
Additionally, after applying the data alignment strategy, the lightweight network GaitBase achieved a Rank-1 accuracy of 74.3\%,  which is only slightly lower than state-of-the-art deep feature extraction network DeepGaitv2-3D-H (74.3\% vs. 75.8\%). 
However, GaitBase has fewer than one-ninth the number of parameters compared to DeepGaitv-3D-H (4.9M vs. 44.4M). 
This indicates that while deep networks require substantial resources to learn data distribution characteristics, these characteristics can be effectively captured using the proposed data alignment strategy.
Furthermore, our method also significantly improved the performance of deep networks, with a 5.4\% increase in Rank-1 accuracy (80.4\% vs. 75.0\%) for DeepGait-P3D-H, thus achieving the state-of-the-art.
In addition, we conducted experiments on multimodal fusion networks, as shown in Table~\ref{tab2}. 
On the multimodal network MSAFF\cite{ShinanZouMSAFF}, our approach achieved a 4.5\% improvement in accuracy(52.6\% vs 48.1\%), demonstrating the effectiveness of our data alignment strategy.

\noindent\textbf{Evaluation of CCPG and SUSTech1K Datasets.} The average Rank-1 accuracy for each network improved by 3.0\% and 2.0\%, respectively. 
Additionally, both datasets achieved the current State-of-the-Art(SOTA) Rank1 accuracy, with CCPG on DeepGaitv2-P3D-H and SUSTech1K on DeepGaitv2-3D-H. For the SUSTech1K dataset, the overall accuracy improvement was relatively modest, which we attribute to the fact that the poses of subjects in this dataset are relatively centered. 

\noindent In summary, the proposed data alignment strategy effectively enhances the performance of gait recognition models, particularly in complex, real-world environments. 


\begin{table}[t]
    \footnotesize
    \caption{Performance comparison of the data alignment strategy in cross-domain tasks. The comparison is made using GaitBase as the baseline network.}
    \label{tab3}
    \centering
    \begin{tabular}{cc|cccc}
        \toprule
        \multicolumn{2}{c|}{Dataset} & \multirow{2}{*}{Rank1} & \multirow{2}{*}{Rank-5} & \multirow{2}{*}{mAP} & \multirow{2}{*}{mINP} \\ \cline{1-2}
        Train         & Tset         &                        &                         &                      &                       \\ \hline
        CCPG\cite{CCPG}          & Gait3D\cite{Gait3D}       & 11.8                   & 22.6                    & 7.8                  & 4.3                   \\
        \rowcolor{gray!10}CCPG-Align    & Gait3D-Align & \textbf{35.8}$\mathrm{\tiny{\hi{+24.0}}}$          & \textbf{52.8}$\mathrm{\tiny{\hi{+30.2}}}$           & \textbf{25.1}$\mathrm{\tiny{\hi{+17.3}}}$        & \textbf{13.7}$\mathrm{\tiny{\hi{+9.4}}}$         \\
        SUSTech1K\cite{SUSTech1K}       & Gait3D       & 21.0                   & 35.1                    & 14.2                 & 8.0                   \\
        \rowcolor{gray!10}SUSTech1K-Align & Gait3D-Align & \textbf{32.6}$\mathrm{\tiny{\hi{+11.6}}}$          & \textbf{48.6}$\mathrm{\tiny{\hi{+13.5}}}$           & \textbf{22.8}$\mathrm{\tiny{\hi{+8.6}}}$        & \textbf{12.9}$\mathrm{\tiny{\hi{+4.9}}}$         \\
        Gait3D        & CCPG         & 38.2                   & -                       & 24.5                 & 8.3                   \\
        \rowcolor{gray!10}Gait3D-Align  & CCPG-Align   & \textbf{45.3}$\mathrm{\tiny{\hi{+7.1}}}$          & -                       & \textbf{31.9}$\mathrm{\tiny{\hi{+7.4}}}$        & \textbf{13.3}$\mathrm{\tiny{\hi{+5.0}}}$                  \\
        \bottomrule    
    \end{tabular}
\end{table}

\begin{table}[t]
    \footnotesize
    \caption{Comparison of the impact of different data alignment strategies on gait recognition performance.}
    \label{tab4}
    \centering
    \begin{tabular}{c|cccc}
        \toprule
        Method                   & Rank-1 & Rank-5 & mAP  & mINP \\ \hline
        w/o Data Alignment       & 64.6   & 80.0   & 54.5 & 36.2 \\
        Random Rotation          & 62.1   & 78.2   & 52.6 & 34.6 \\
        Minimum Bounding Box     & 69.8   & 83.7   & 60.3 & 41.1 \\
        Skeleton-guided Rotation & \textbf{74.3}   & \textbf{87.9}   & \textbf{66.2} & \textbf{46.9} \\
        \bottomrule    
    \end{tabular}
\end{table}

\subsection{Ablation Study}
In this section, we performed cross-dataset experiments to validate the generalizability of the data alignment strategy. Furthermore, we compared various alignment approaches to highlight the superior performance of skeleton-based methods.

\noindent\textbf{Cross-Domain Evaluation.}
As illustrated in Table~\ref{tab3}, we compare the performance of data alignment strategies in the cross-dataset task. 
We use GaitBase as the baseline and evaluate its performance across the Gait3D, CCPG, and SUSTech1K datasets. 
The results demonstrate that networks trained with data alignment shows stronger adaptability across different datasets.
When training and validating on the origin CCPG and Gait3D datasets, the Rank-1 accuracy was only 11.8\%. 
The low accuracy can be attributed to the significant distribution discrepancies between the training and validation datasets. 
After applying data alignment, the accuracy improved by 24.0\% (35.8\% vs. 11.8\%), indicating that data alignment helps the network extract more general gait features by reducing inconsistencies such as variations in viewpoint and posture. 
When pedestrians maintain spatio-temporal distribution consistency across different frames, the network is better able to focus on critical variations rather than noise, thereby exhibiting superior generalization capability.
Moreover, a similar trend was observed on the SUSTech1K dataset. 
After data alignment, the Rank-1 accuracy improved from 21.0\% to 32.6\%, showing an improvement of 11.6\%.

\noindent\textbf{Impact of Alignment Strategies.}
In addition to comparing the performance of various alignment methods, we also investigate the effects of several alternative data alignment strategies. 
Inspired by the work of Wang et al. \cite{QAGait}, we consider two additional alignment strategies: restricted random rotation and minimum bounding box rotation. 
The restricted random rotation strategy determines the rotation direction by analyzing the relative proportion of the foreground parts on the left and right sides of the silhouette image, then applies a randomly rotation angle. 
For instance, when the person is biased toward the left side of the image, the strategy rotates the silhouette to the right. The minimum bounding box rotation method first identifies the minimal bounding box that encloses the foreground region of the image and then rotates the bounding box to align its height vertically.
As shown in the experimental results in Table~\ref{tab4}, the restricted random rotation strategy reduces accuracy due to the unpredictability introduced by random rotation angles. In contrast, the minimum bounding box rotation strategy improves accuracy by 5.2\% (69.8\% vs. 64.6\%). However, our proposed skeleton-guided data alignment method outperforms all other strategies, achieving a significant accuracy improvement of 9.7\%.

\section{CONCLUSION}
In summary, we investigated the underlying causes of the performance degradation observed in existing gait recognition methods when transitioning from laboratory datasets to wild datasets. Our analysis highlights the critical role of data alignment in addressing the challenges associated with wild datasets. To this end, we proposed a data-alignment-based gait recognition framework \textbf{DAGait}, designed to enhance the spatio-temporal consistency of training data. Experimental results demonstrate that our alignment strategy leads to significant accuracy improvements across multiple datasets and methods, highlighting its effectiveness and generalizability.

\section*{ACKNOWLEDGMENTS}
This research was funded through National Key Research and Development Program of China (Project No. 2022YFB36066), in part by the Shenzhen Science and Technology Project under Grant (KJZD20240903103210014)

\section*{A\quad SUPPLEMENTARY MATERIAL}
This supplementary material first presents additional experiments on DeepGait with different parameter scales, followed by a comprehensive cross-dataset evaluation.

\section*{A.1\quad Additional Experiments on DeepGait}
We conducted data alignment experiments on all networks in the DeepGait\cite{deepgait} series, as shown in Table~\ref{tab6}. After applying data alignment, the Rank-1 accuracy on the Gait3D, CCPG, and SuTech1K\cite{SUSTech1K} datasets improved by an average of 4.95\%, 2.3\%, and 2.0\%, respectively. Specifically, for the Gait3D dataset, our method achieved a 7.7\% increase in Rank-1 accuracy on the DeepGait3D-B model (72.2\% vs. 64.5\%). Furthermore, on the DeepGaitP3D-H model, the Rank-1 accuracy reached 80.4\%, establishing a new state-of-the-art (SOTA) performance for this challenging dataset.These results validate the effectiveness of our proposed data alignment strategy.

\begin{table*}[ht]
    \footnotesize
    \centering
    \caption{Performance comparison of the original Gait3D, CCPG, and SUSTech1K datasets versus their data-aligned counterparts, evaluated on the DeepGait network.}        
    \label{tab6}
    \begin{tabular}{cl|cccccccccc}
        \toprule
            \multicolumn{2}{c|}{\multirow{4}{*}{Method}}  & \multicolumn{10}{c}{Testing Datasets}                                                                                                                                                                     \\ \cline{3-12} 
            \multicolumn{2}{c|}{}                         & \multicolumn{3}{c|}{Gait3D}                                                                   & \multicolumn{5}{c|}{CCPG}                             & \multicolumn{2}{c}{SUSTech1K}                     \\ \cline{3-12} 
            \multicolumn{2}{c|}{}                         & \multirow{2}{*}{Rank-1} & \multirow{2}{*}{Rank-5} & \multicolumn{1}{c|}{\multirow{2}{*}{mAP}} & CL   & UP   & DN   & BG   & \multicolumn{1}{c|}{Mean} & \multirow{2}{*}{Rank-1} & \multirow{2}{*}{Rank-5} \\ \cline{6-10}
            \multicolumn{2}{c|}{}                         &                         &                         & \multicolumn{1}{c|}{}                     & \multicolumn{5}{c|}{Rank-1}                           &                         &                         \\ \hline
            \multicolumn{2}{c|}{DGaitv2-2D-B(Arxiv2023)\cite{deepgait}}  & 64.5                    & 81.7                    & \multicolumn{1}{c|}{56.5}                 & 69.5 & 75.6 & 76.9 & 84.5 & \multicolumn{1}{c|}{76.6} & 69.7                    & 86.5                    \\
            \rowcolor{gray!10} \multicolumn{2}{c|}{DGaitv2-2D-B+Ours}                     & 72.2$\mathrm{\tiny{\hi{+7.7}}}$                    & 86.9$\mathrm{\tiny{\hi{+5.2}}}$                    & \multicolumn{1}{c|}{63.7$\mathrm{\tiny{\hi{+7.2}}}$}                 & 73.5$\mathrm{\tiny{\hi{+4.0}}}$ & 78.4$\mathrm{\tiny{\hi{+2.8}}}$ & 80.7$\mathrm{\tiny{\hi{+3.8}}}$ & 85.8$\mathrm{\tiny{\hi{+1.3}}}$ & \multicolumn{1}{c|}{79.6$\mathrm{\tiny{\hi{+3.0}}}$} & 72.0$\mathrm{\tiny{\hi{+2.3}}}$                    & 87.4$\mathrm{\tiny{\hi{+0.9}}}$                    \\ 
            \multicolumn{2}{c|}{DGaitv2-2D-L(Arxiv2023)\cite{deepgait}}  & 67.8                    & 83.9                    & \multicolumn{1}{c|}{59.7}                 & 69.9 & 76.1 & 77.1 & 84.4 & \multicolumn{1}{c|}{76.9} & 74.8                    & 89.2                    \\
            \rowcolor{gray!10} \multicolumn{2}{c|}{DGaitv2-2D-L+Ours}                     & 75.2$\mathrm{\tiny{\hi{+7.4}}}$                    & 88.4$\mathrm{\tiny{\hi{+4.5}}}$                    & \multicolumn{1}{c|}{67.2$\mathrm{\tiny{\hi{+7.5}}}$}                 & 73.8$\mathrm{\tiny{\hi{+3.9}}}$ & 78.0$\mathrm{\tiny{\hi{+1.9}}}$ & 81.4$\mathrm{\tiny{\hi{+4.3}}}$ & 86.2$\mathrm{\tiny{\hi{+1.8}}}$ & \multicolumn{1}{c|}{80.0$\mathrm{\tiny{\hi{+3.1}}}$} & 77.3$\mathrm{\tiny{\hi{+2.5}}}$                    & 90.2$\mathrm{\tiny{\hi{+1.0}}}$                    \\
            \multicolumn{2}{c|}{DGaitv2-3D-B(Arxiv2023)\cite{deepgait}}  & 71.0                    & 85.0                    & \multicolumn{1}{c|}{62.3}                 & 78.4 & 82.8 & 80.9 & 88.3 & \multicolumn{1}{c|}{82.6} & 77.3                    & 90.3                    \\
            \rowcolor{gray!10} \multicolumn{2}{c|}{DGaitv2-3D-B+Ours}                     & 74.9$\mathrm{\tiny{\hi{+3.9}}}$                    & 88.5$\mathrm{\tiny{\hi{+3.5}}}$                    & \multicolumn{1}{c|}{67.6$\mathrm{\tiny{\hi{+5.3}}}$}                 & 80.1$\mathrm{\tiny{\hi{+1.7}}}$ & 84.7$\mathrm{\tiny{\hi{+1.9}}}$ & 84.1$\mathrm{\tiny{\hi{+3.2}}}$ & 89.0$\mathrm{\tiny{\hi{+0.7}}}$ & \multicolumn{1}{c|}{84.5$\mathrm{\tiny{\hi{+1.9}}}$} & 78.4$\mathrm{\tiny{\hi{+1.1}}}$                    & 90.4$\mathrm{\tiny{\hi{+0.1}}}$                    \\ 
            \multicolumn{2}{c|}{DGaitv2-3D-L(Arxiv2023)\cite{deepgait}}  & 74.1                    & 87.0                    & \multicolumn{1}{c|}{66.5}                 & 77.9 & 83.2 & 79.9 & 88.6 & \multicolumn{1}{c|}{82.4} & 79.4                    & 91.2                    \\
            \rowcolor{gray!10} \multicolumn{2}{c|}{DGaitv2-3D-L+Ours}                     & 78.5$\mathrm{\tiny{\hi{+4.4}}}$                    & 89.7$\mathrm{\tiny{\hi{+2.7}}}$                    & \multicolumn{1}{c|}{71.0$\mathrm{\tiny{\hi{+4.5}}}$}                 & 79.6$\mathrm{\tiny{\hi{+1.7}}}$ & 85.0$\mathrm{\tiny{\hi{+1.8}}}$ & \textbf{84.1}$\mathrm{\tiny{\hi{+4.2}}}$ & 88.7$\mathrm{\tiny{\hi{+0.1}}}$ & \multicolumn{1}{c|}{84.4$\mathrm{\tiny{\hi{+2.0}}}$} & 81.5$\mathrm{\tiny{\hi{+2.1}}}$                    & 92.1$\mathrm{\tiny{\hi{+0.9}}}$                    \\
            \multicolumn{2}{c|}{DGaitv2-3D-H(Arxiv2023)\cite{deepgait}}  & 75.8                    & 87.3                       & \multicolumn{1}{c|}{67.3}                    & 75.9 & 83.2 & 79.1 & 88.7 & \multicolumn{1}{c|}{81.8} & 80.4                    & 91.6                    \\
            \rowcolor{gray!10} \multicolumn{2}{c|}{DGaitv2-3D-H+Ours}                     & 79.4$\mathrm{\tiny{\hi{+3.6}}}$                    & 89.8$\mathrm{\tiny{\hi{+2.5}}}$                    & \multicolumn{1}{c|}{72.7$\mathrm{\tiny{\hi{+5.4}}}$}                 & 78.0$\mathrm{\tiny{\hi{+2.1}}}$ & 83.8$\mathrm{\tiny{\hi{+0.6}}}$ & 83.2$\mathrm{\tiny{\hi{+4.1}}}$ & 89.0$\mathrm{\tiny{\hi{+0.3}}}$ & \multicolumn{1}{c|}{83.5$\mathrm{\tiny{\hi{+1.7}}}$} & \textbf{82.8}$\mathrm{\tiny{\hi{+2.4}}}$                    & 92.7$\mathrm{\tiny{\hi{+1.1}}}$                    \\
            \multicolumn{2}{c|}{DGaitv2-P3D-B(Arxiv2023)\cite{deepgait}}  & 70.8                    & 85.7                       & \multicolumn{1}{c|}{62.9}                    & 76.0 & 81.6 & 80.4 & 87.8 & \multicolumn{1}{c|}{81.5} & 76.5                    & 90.7                    \\
            \rowcolor{gray!10} \multicolumn{2}{c|}{DGaitv2-P3D-B+Ours}                     & 73.5$\mathrm{\tiny{\hi{+2.7}}}$                    & 86.9$\mathrm{\tiny{\hi{+1.2}}}$                    & \multicolumn{1}{c|}{65.6$\mathrm{\tiny{\hi{+2.7}}}$}                 & 79.3$\mathrm{\tiny{\hi{+3.3}}}$ & 82.9$\mathrm{\tiny{\hi{+1.3}}}$ & 83.6$\mathrm{\tiny{\hi{+3.2}}}$ & 88.2$\mathrm{\tiny{\hi{+0.4}}}$ & \multicolumn{1}{c|}{83.5$\mathrm{\tiny{\hi{+2.0}}}$} & 78.1$\mathrm{\tiny{\hi{+1.6}}}$                    & 91.9$\mathrm{\tiny{\hi{+1.2}}}$                    \\
            \multicolumn{2}{c|}{DGaitv2-P3D-L(Arxiv2023)\cite{deepgait}} & 74.2                    & 86.9                    & \multicolumn{1}{c|}{67.1}                 & 76.8 & 82.5 & 79.8 & 88.2 & \multicolumn{1}{c|}{81.8} & 79.6                    & 91.4                    \\
            \rowcolor{gray!10} \multicolumn{2}{c|}{DGaitv2-P3D-L+Ours}                     & 78.9$\mathrm{\tiny{\hi{+4.7}}}$                    & 89.8$\mathrm{\tiny{\hi{+2.9}}}$                    & \multicolumn{1}{c|}{71.5$\mathrm{\tiny{\hi{+4.4}}}$}                 & 79.9$\mathrm{\tiny{\hi{+3.1}}}$ & 84.0$\mathrm{\tiny{\hi{+1.5}}}$ & 83.8$\mathrm{\tiny{\hi{+4.0}}}$ & 89.0$\mathrm{\tiny{\hi{+0.8}}}$ & \multicolumn{1}{c|}{84.3$\mathrm{\tiny{\hi{+2.5}}}$} & 81.6$\mathrm{\tiny{\hi{+2.0}}}$                    & 92.3$\mathrm{\tiny{\hi{+0.9}}}$                    \\
            \multicolumn{2}{c|}{DGaitv2-P3D-H(Arxiv2023)\cite{deepgait}} & 75.0                    & 87.3                       & \multicolumn{1}{c|}{66.9}                    & 76.2 & 83.5 & 80.0 & 89.4 & \multicolumn{1}{c|}{82.3} & 80.7                    & 91.7                    \\
            \rowcolor{gray!10} \multicolumn{2}{c|}{DGaitv2-P3D-H+Ours}                     & \textbf{80.4}$\mathrm{\tiny{\hi{+5.4}}}$                    & \textbf{90.4}$\mathrm{\tiny{\hi{+3.1}}}$                    & \multicolumn{1}{c|}{\textbf{72.9}$\mathrm{\tiny{\hi{+6.0}}}$}                 & \textbf{80.5}$\mathrm{\tiny{\hi{+4.3}}}$ & \textbf{86.0}$\mathrm{\tiny{\hi{+2.5}}}$ & 83.8$\mathrm{\tiny{\hi{+3.8}}}$ & \textbf{89.6}$\mathrm{\tiny{\hi{+0.2}}}$ & \multicolumn{1}{c|}{\textbf{85.0}$\mathrm{\tiny{\hi{+2.7}}}$} & 82.6$\mathrm{\tiny{\hi{+1.9}}}$                    & \textbf{92.9}$\mathrm{\tiny{\hi{+1.2}}}$                    \\         

        \bottomrule 
    \end{tabular}
\end{table*}

\section*{A.2\quad Detailed Cross-Dataset Evaluation}
As illustrated in Table~\ref{tab7}, we compare the performance of data alignment strategies in the cross-dataset task. We use GaitBase\cite{OpenGait} as the baseline network, Gait3D and CCPG representing the original datasets, while Gait3D-Align and CCPG-Align refer to the datasets processed with data alignment. In the first scenario, neither of the datasets undergoes any form of alignment. Under these conditions, the model’s prediction accuracy is relatively low due to the significant domain gap between Gait3D and CCPG. When the CCPG dataset is aligned, we observe a significant reduction in spatiotemporal inconsistencies in the test set. However, since the training set remains unaligned, the gait features learned by the model still contain domain-specific biases from Gait3D, resulting in only a modest accuracy improvement (38.9\% vs. 38.2\%).
When the model is trained on the aligned Gait3D dataset while leaving the test set unchanged, there is a marked increase in accuracy (41.9\% vs. 38.2\%). This improvement suggests that training on an aligned dataset enhances the model's robustness, enabling it to learn more standardized and stable gait features. When data alignment is applied in both the training and testing phases, the model’s accuracy increases by 7.1\% (45.3\% vs. 38.2\%). This dual-stage alignment significantly boosts the model's ability to adapt to domain discrepancies, thereby maximizing feature transferability and robustness.

\begin{table}[ht]
    \footnotesize
    \caption{Performance comparison of the data alignment strategy in cross-dataset tasks. The comparison is made using GaitBase as the baseline network.}
    \label{tab7}
    \centering
    \begin{tabular}{cc|ccccc}
        \toprule
        \multicolumn{2}{c|}{Dataset} & \multicolumn{5}{c}{Rank-1}       \\ \hline
        Train          & Test        & CL   & UP   & DN   & BG   & Mean \\ \hline
        Gait3D         & CCPG        & 13.8 & 23.8 & 45.8 & 69.6 & 38.2 \\
        Gait3D         & CCPG-Align  & 13.5 & 23.5 & 46.8 & 71.7 & 38.9 \\
        Gait3D-Align   & CCPG        & 16.1 & 26.9 & 49.2 & 75.2 & 41.9 \\
        Gait3D-Align   & CCPG-Align  & \textbf{17.5} & \textbf{27.9} & \textbf{55.6} & \textbf{80.1} & \textbf{45.3} \\ 
        \bottomrule    
    \end{tabular}
\end{table}

\bibliographystyle{IEEEbib}
\bibliography{icme2025references}

\vspace{12pt}

\end{document}